\documentclass[runningheads]{llncs}
\usepackage{cite}
\usepackage{derivative}
\usepackage{amsmath}
\usepackage{amssymb}
\usepackage{amstext}
\usepackage[utf8]{inputenc}
\usepackage[ruled,longend]{algorithm2e}
\usepackage{textgreek}
\usepackage{algpseudocode}
\usepackage{subcaption,graphicx}

\newcommand{\Lagr}{\mathcal{L}}
\usepackage{graphicx}
\begin{document}
\title{Exploring Action-Centric Representations Through the Lens of Rate-Distortion Theory}

\titlerunning{Action-Centric Representations and Rate-Distortion Theory}

%
%
\author{Miguel De Llanza Varona\inst{1,2}, Christopher Buckley\inst{1,2}, Beren Millidge\inst{3}}
%
%
\institute{School of Engineering and Informatics, University of
Sussex, Brighton, UK \and VERSES Research Lab, Los Angeles, California, USA \email{M.De-Llanza-Varona@sussex.ac.uk, C.L.Buckley@sussex.ac.uk} \and MRC Brain Networks Dynamics Unit, University of Oxford, Oxford, UK \email{beren@millidge.name}}
\maketitle              
\begin{abstract}
Organisms have to keep track of the information in the environment that is relevant for adaptive behaviour. Transmitting information in an economical and efficient way becomes crucial for limited-resourced agents living in high-dimensional environments. The efficient coding hypothesis claims that organisms seek to maximize the information about the sensory input in an efficient manner. Under Bayesian inference, this means that the role of the brain is to efficiently allocate resources in order to make predictions about the hidden states that cause sensory data. However, neither of those frameworks accounts for how that information is exploited downstream, leaving aside the action-oriented role of the perceptual system. Rate-distortion theory, which defines optimal lossy compression under constraints, has gained attention as a formal framework to explore goal-oriented efficient coding. In this work, we explore action-centric representations in the context of rate-distortion theory. We also provide a mathematical definition of abstractions and we argue that, as a summary of the relevant details, they can be used to fix the content of action-centric representations. We model action-centric representations using VAEs and we find that such representations i) are efficient lossy compressions of the data; ii) capture the task-dependent invariances necessary to achieve successful behaviour; and iii) are not in service of reconstructing the data. Thus, we conclude that full reconstruction of the data is rarely needed to achieve optimal behaviour, consistent with a teleological approach to perception.

\keywords{Rate-distortion theory \and Action-centric representations \and Efficient coding \and Bayesian Inference} 
\end{abstract}

\raggedbottom
\section{Introduction}
Embodied agents have to focus on the relevant information from their environment to achieve adaptive behaviour. Their resource-limited cognition and the high-complexity structure inherent to the environment force them to economize the transmission of information. Thus, the goal of the perceptual system is to generate representations that are useful for successful behaviour while at the same being encoded in the most efficient manner.  

A well-known hypothesis in theoretical neuroscience called the efficient coding hypothesis proposes that the neural coding in the brain is optimized to maximize sensory information under metabolic and capacity constraints \cite{Laughlin_neuron_capac,barlow_sensory,brain_economic}. In particular, this hypothesis suggests that neurons are tuned to the statistical properties of the environment, which allows them to efficiently allocate signaling resources to generate compressed low-dimensional representations of the environment. In this theoretical framework, it is commonly assumed that the function of neurons is to maximize their capacity to account for all the variability in the sensory input. In information theory terms, this means that the brain seeks to maximize the mutual information between stimuli and neurons' response to reduce as much as possible the uncertainty about the environment, which is defined by its entropy. While this hypothesis answers the question about information processing under biological constraints, it leaves aside the utilitarian aspect of perception \cite{sims_compress,Griffiths,bayesian_coding,efficient_cod_review,Genewein}. 

Cognition can't be fully understood without its ecological context, as agents are coupled with their environment forming a perception-action feedback loop \cite{deWit}. In this sense, the functional role of perceptual processing has to be in service of achieving behavioural objectives, and to do that, perceptual representations must efficiently encode the relevant information needed by the motor system to guide future actions. Thus, a key component of the perceptual system is to summarize relevant sensory information to  generate action-centric representations.

The teleological essence of the perceptual system imposes a normativity on representations: a perceptual representation is accurate if it captures the relevant information needed downstream and discards the irrelevant details. Thus, we need an extra ingredient to account for the goodness of representations under constraints. This is where the rate-distortion theory comes into play \cite{Shannon_RD}. This subfield of information theory defines the optimal trade-off between channel capacity and expected communication error. When error-free communication is not necessary to guide behaviour, the optimal encoding is a lossy compression of the input. 

Interestingly, rate-distortion theory can be seen as a way to perform Bayesian inference under constraints. Under a Bayesian approach to cognition, the brain performs inference to compute an optimal posterior distribution over hidden environmental states given sensory data \cite{karl_guide}. As computing the true posterior is usually intractable, the brain approximates the true posterior by optimizing the variational free energy \cite{karl_fep,karl_fep_unified,buckley_fep}. The main conceptual contribution of rate-distortion theory is to define the “goodness” of that approximation, as computing the true posterior is not always necessary to act optimally. In the context of active inference, it has been shown that action-oriented models learn parsimonious representations of the environment by capturing relevant information for behaviour \cite{action_oriented}. In the same spirit, we investigate the information-theoretic properties of action-centric representations and their relation to the formal definition of abstractions we propose.

In this work, we explore action-centric representations under the lens of rate-distortion theory to account for the teleological aspect of perception. To do that, we provide a mathematical definition of abstraction that allows us to specify the task-relevant information that should carry an action-centric representation. Given the tight connection between Bayesian Inference and rate-distortion theory, we use a Variational Autoencoder (VAE) framework to model action-centric representations as optimal lossy compression. Our results show that action-centric representations are optimal lossy compressions of the data; can be successfully used in downstream tasks; and crucially, they achieve that without being in service of reconstructing the data.

\section{Efficient coding and rate-distortion theory}
\subsection{Efficient coding}
The efficient coding hypothesis states that neurons are optimized to maximize the information they carry about sensory states. In doing so, neurons have to generate minimal redundancy codes to economically use limited resources. In particular, neurons seek to maximize the ratio between information about sensory inputs, defined by the mutual information $I(X;Z)$ between sensory data $X$ and neural responses $Z$, and the channel capacity $C$: $\frac{I(X;Z)}{C}$. The maximum mutual information is upper bounded by the channel capacity
\begin{align}
    C \geq I(X;Z)
\end{align}
so the best efficient coding satisfies
\begin{align}
    I(X;Z) = C
\end{align}
where neuronal encoding exploits the whole bandwidth of the channel.

\subsection{Rate-distortion theory as goal-oriented efficient coding}
\begin{figure}
    \centering \includegraphics[width=0.5\textwidth]{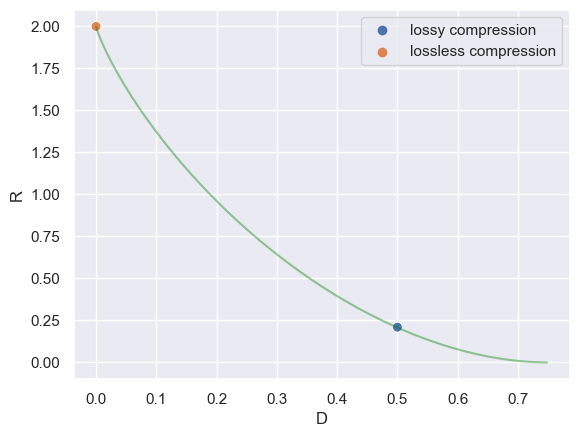}
    \caption{Rate-distortion function for a discrete random variable with four uniformly distributed states. Assuming that behavioral objectives are achievable even when half of the information generated at the source is missing; that is, when the expected distortion $D$ does not exceed 0.5 (x-axis), an optimal agent with bounded rationality can rely on a lossy compression scheme and transmit information at a rate $R$ of 0.20 bits (y-axis).}\label{fig:loss_lossless}
    \end{figure}

Under the classical conception of efficient coding, the exploitation of information downstream is ignored. When not all sensory information is needed to guide behaviour, error-free communication is not expected. This is precisely what is addressed by the rate-distortion theory, which provides the theoretical foundations for optimal lossy data compression. Formally, the rate-distortion function defines an optimal lossy compression $Z$ of some data $X$ as the minimization of their mutual information $I(X;Z)$ given some expected distortion $D$ associated with reconstructing $X$ from its lossy compression $Z$. It is defined as\cite{Cover_Thomas}
\begin{align}
    R(D) = \min_{q(z|x): D_{q(x,z)} \leq D}  I(X;Z)
\end{align}
where $q$ is the optimal distribution of $z$ given $x$ that satisfies the expected distortion constraint and the rate $R$ is an upper bound on the mutual information:
\begin{align}\label{rate_mutual_info}
R \geq I(X;Z)
\end{align}
The expected distortion $D$ is defined by some arbitrary loss function (e.g., mean-squared error) that quantifies the faithfulness of information transmission (i.e., how well can the data be recovered from its optimal lossy compression). Lossy compression
sacrifices the capacity to represent all the information in the input in service of transmitting information that allows adaptive behaviour. Having a faithfulness criterion allows the brain to efficiently represent the world by allocating just the necessary amount of resources required to navigate the environment (Figure \ref{fig:loss_lossless}). Thus, rate-distortion adds a teleological perspective to efficient coding that shifts the focus from efficient information maximization to efficient transmission of action-oriented information.

In the lossy regime of the rate-distortion (i.e., all points such that $D>0$), the obtained representations can be understood as abstractions of the data, as their function is to summarize the relevant properties of the data needed downstream. In the next section, we provide a mathematical definition of abstractions based on the intuition that are entities that convey the necessary information to answer a set of queries about the data. The mathematical formulation of abstractions is crucial to determine the content of action-centric representations.

\section{Abstractions and action-centric representations}
\subsection{Mathematical formalization of abstractions}
An abstraction is the reduction of complexity by discarding certain features while preserving others. As a relational concept, an abstraction involves two components: its object (what is being abstracted) and its content (what the abstraction is about). The content of the abstraction is a summary of the \textit{relevant} properties of its object and the relevancy is fixed, ultimately, by the agent's needs. Thus, abstractions are intrinsically \textit{teleological} entities; that is, their meaning or content is fixed by their function or purpose, which is to transmit information about the properties of interest for an agent.

Following \cite{Beren_Abs}, we address the content of abstractions as the information necessary to answer a set of queries about the data. A query captures what the agent wants to know about the data (i.e., what is relevant). Formally, given a set of queries about the data $\mathbb{Q} = \{Q_1, Q_2\dots Q_3\}$, each query is a mapping from the data distribution to a probability distribution over a subset of elements of the data $Q:\mathbb{X} \rightarrow p(q|x)$. A good abstraction is one that fulfills its purpose; namely, one that keeps track of those properties that make it possible to answer a particular query. Thus, the `goodness' of an abstraction $z$ for a given query can be defined as:
\begin{align}\label{goodness_abstraction}
    \Lagr_Q(x,z) = \mathcal{D}[Q(x)||Q(r(z))]
\end{align}
where $Q(x)$ is the query distribution
over the true system or data, $Q(r(z))$ is the query distribution over a lossy reconstruction of the data $r(z)$ produced by the abstraction $z$, and $\mathcal{D}$ is an arbitrary divergence function.  
Without loss of generality, the `goodness' of an abstraction given a set of queries can be defined as the weighted loss over all the queries given the abstraction:
\begin{align}\label{goodness_avg_abstraction}
    \Lagr(x,z) = \sum_{Q_i\in\mathbb{Q}}p(Q_i)\Lagr_{Q_i}(x,z)
\end{align}

Ideally, the mutual information between the abstractions and the query should be the same as the information transmitted between the data and the query:
\begin{align}\label{MI}
    I(X;Q) = I(Q;Z)
\end{align}
The intuition is that a good abstraction $Z$ of the data should reduce the uncertainty of the data $X$ in the same way as the query $Q$ does.

\subsection{Abstractions as sufficient and non-superfluous representations}
Following \cite{robust_repr}, an abstract representation $Z$ that captures the relevant details of the data $X$ to answer a query $Q$ should be sufficient ($I(X;Q|Z) = 0$) and non-superfluous ($I(X;Z|Q) = 0$):
\begin{align}
\underbrace{{I(X;Z|Q) }}_{\text{Superfluous}} &= I(X;Z) - I(X;Q;Z)\\ 
    &= I(X;Z) - I(X;Q) + \underbrace{{I(X;Q|Z)}}_{\text{Sufficient}}\\
    &= I(X;Z) - I(X;Q)
\end{align}
therefore
\begin{align}
    \big( I(X;Z|Q) = 0 \big) \land \big( I(X;Q|Z) = 0 \big) \iff I(X;Z) = I(X;Q)
\end{align}
From an information theory perspective, a good abstraction $Z$ only carries the relevant information in the data; that is, the information necessary to answer a query.
Note that this is a continuum where at one extreme the optimal compression captures all the information in the data when the query contains the same information as the data $H(Q) = H(X)$ (an ideal scenario in the efficient coding hypothesis, where the goal is to maximize mutual information):
\begin{align}
    I(X;Q) = H(X) - H(X|Q) = H(X) - H(X|X) = H(X)
\end{align}
therefore
\begin{align}
    I(X;Z) &= I(X;Q)\\ 
    I(X;Z) &= H(X)
\end{align}
which corresponds to the lossless compression regime of the rate-distortion function. On the contrary, when the communication channel is closed, then we recover the other extreme of the rate-distortion curve, where the mutual information is zero. This is the case when knowing the query does not reduce the uncertainty of the data:
\begin{align}
    I(X;Q) = H(X) - H(X|Q) = H(X) - H(X) = 0
\end{align}
therefore
\begin{align}
    I(X;Z) &= I(X;Q)\\ 
    I(X;Z) &= 0
\end{align}
Any other stage in between is a case where the query carries partial information about the data. Importantly, these information-theoretic entities are implicitly optimized in rate-distortion theory. On the one hand, sufficient information is related to predictability and, therefore, to communication fidelity, which is satisfied when the expected distortion allows for answering the query (i.e., successful behaviour). On the other hand, non-superfluous information is related to the minimization of the mutual information up to a point in which only query-relevant information is encoded in the abstraction. Thus, optimal lossy representations, whose function is to encode the relevant invariances and symmetries in the data, lie in the rate-distortion curve. 

\section{Variational Free Energy and Rate-distortion theory}
As computing the rate-distortion function is intractable in high-dimensional systems \cite{Cover_Thomas}, variational inference can be used as a proxy of the amount of information transmitted through a communication channel. In variational inference, a quantity called variational free energy sets an upper bound on the sensory surprisal (i.e., the entropy of sensory states), and by minimizing it is reduced the uncertainty about the sensory data allowing for predictability of future states and adaptive behaviour. One common variational free energy decomposition is the ELBO, which involves two terms, accuracy and complexity, and is formally defined as\cite{pred_coding_review}
\begin{align}
    F &= \int q(z|x) \ln{\frac{q(z|x)}{p(x,z)}}\\
    F &= \int q(z|x) \ln{\frac{q(z|x)}{p(z)}} - \int q(z|x) \ln{p(x|z)}\\
    F &= {\underbrace{{D_{KL}[q(z|x) || p(z)]}}_{\text{Complexity}}} - {\underbrace{{\mathop{\mathbb{E}}_{z \sim q}[\ln{p(x|z)}] }}_{\text{Accuracy}}}\label{elbo}
\end{align}

To model lossy representations of the data, we use  VAEs due to its close relation to variational inference and rate-distortion theory. VAEs is an  unsupervised learning framework that captures the underlying data distribution by using i) an encoder that learns a latent representation of the data; and ii) a decoder that generates data-like samples from the latent representation. The objective function optimized by VAEs is the ELBO, where the complexity term can be seen as a regularizer applied to the latent space, and the accuracy term as the faithfulness of the decoder's reconstruction. \\ 

As has been recently shown \cite{Alemi_Fixing_ELBO,Hoffman_ELBO_surgery}, the ELBO is implicitly optimizing the rate-distortion function. On the one hand, the expected complexity is an upper bound on the mutual information $I(X;Z)$ (see Appendix \ref{elbo_mutual} for full derivation):
\begin{align}
    \mathop{\mathbb{E}}_{p(x)} \bigl[ D_{KL}[q(z|x) || p(z)] \bigl] \geq I(X;Z)
\end{align}
just as the rate $R$ is in rate-distortion theory.  
On the other hand, the expected distortion can be measured using any loss function that captures how faithful the reconstruction of the decoder resembles the input data (e.g., hamming distance). In this case, the negative log-likelihood used in VAEs can be used as a distortion measure between the input and its reconstruction, so $D$ can be defined as:
\begin{equation}\label{distortion_measure}
    D = -\mathop{\mathbb{E}}_{z\sim{q_\phi(z|x)}}\big[\log{p_\theta(x|z)}\big]
\end{equation}
Thus, variational inference can be understood through the lens of rate-distortion is characterized as
\begin{equation}\label{Loss_VAEs_RD}
    F = \underbrace{{-\mathop{\mathbb{E}}_{z\sim{q_\phi(z|x)}}\big[\log{p_\theta(x|z)}\big]}}_{\text{Distortion}} + {\underbrace{{\mathcal{D}_{KL}\big[q_\phi(z|x)||p(z)\big]}}_{\text{Rate}}}
\end{equation}

\section{Methods}
\subsection{Model} 
\begin{figure}
    \centering
    \begin{subfigure}[b]{0.5\linewidth}
    \includegraphics[width=1\textwidth,height=3.5cmf]{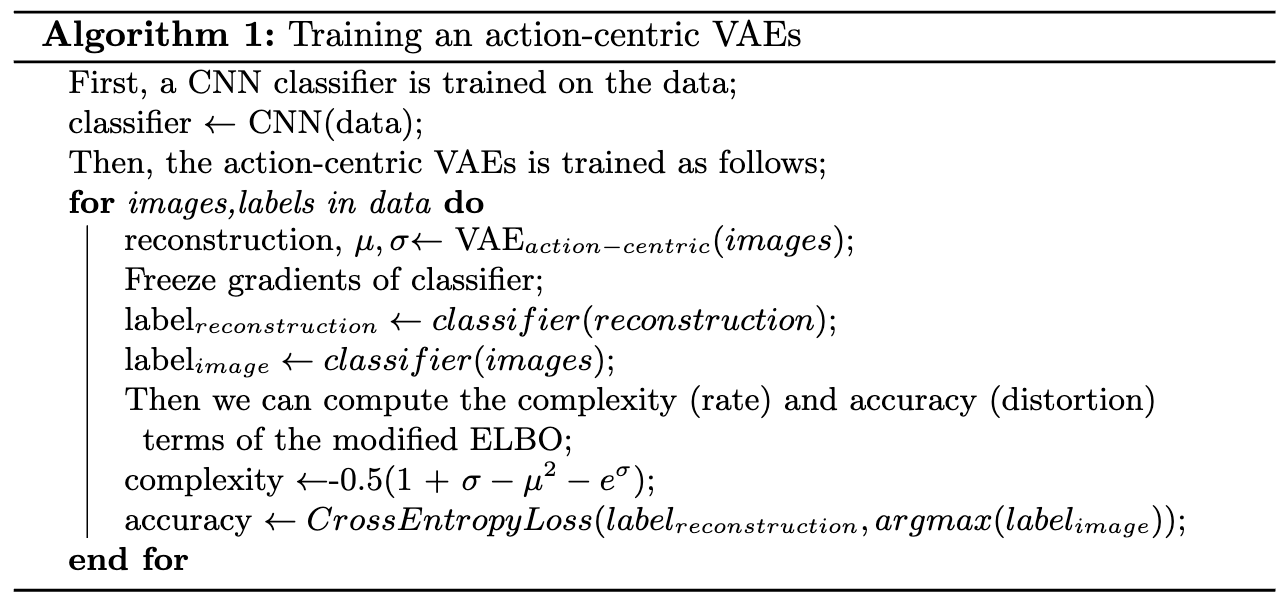}
    \caption{Algorithmic details of the action-centric VAEs}
    \end{subfigure}
    \begin{subfigure}[b]{0.4\linewidth}
    \includegraphics[width=1\textwidth]{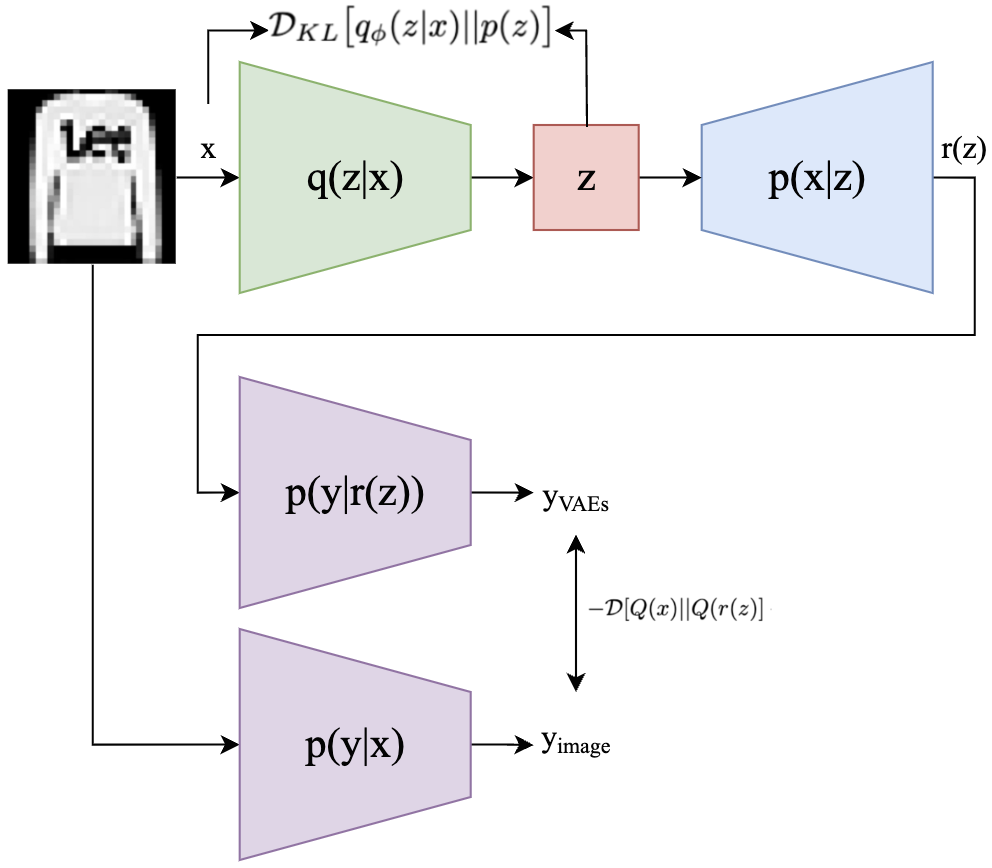}
    \caption{Architecture of the action-centric VAEs}
    \end{subfigure}%
    \caption{Action-centric VAEs. Left: algorithmic level of the action-centric VAEs. Right: a schematic overview of the architecture. The novel component is in the accuracy term of the ELBO. Instead of measuring the faithfulness of the reconstruction, it is measured how good is the reconstruction for a specific task, which in this case is an image classifier.}\label{fig:custom_VAEs}
\end{figure}

Inspired by the utilitarian perspective on the efficient coding hypothesis and the mathematical foundations of abstractions, we present a modified VAEs to model action-centric representations (Figure \ref{fig:custom_VAEs}). The main novelty of the VAEs presented here lies in the accuracy term of the free energy (Eq. \eqref{elbo}). Contrary to vanilla VAEs, where the goal is to learn latent representations of the data to reconstruct it as faithfully as possible, here we are interested in learning action-centric representations that convey sufficient and non-superfluous information about a query. In this model, full reconstruction of the data is not expected. The final form of the objective function for our action-centric VAEs is:
\begin{align}
    \label{loss_custom_vaes}
    F =  -\mathcal{D}[Q(x)||Q(r(z))] + \beta\mathcal{D}_{KL}\big[q_\phi(z|x)||p(z)\big]
\end{align}  
where $\beta$ is the gradient of the rate with respect to the distortion $\pdv{R}{D}=\beta$ and here it's used to target specific regimes of the rate-distortion plane \cite{beta_vae}. The accuracy is modified to account for the goodness of the abstraction.
The training pipeline is as follows. First, we define a query to be the discrimination of the ten classes of the FASHION-MNIST dataset. We first trained a classifier, using a CNN, on the task specified by the query (i.e., multiclass classification). Once the discriminator is trained we trained both the vanilla VAEs and our action-centric VAEs. Importantly, both VAEs have the same channel capacity, as they share the same architecture, so the maximum achievable rate in both models is the same. The crucial difference is that our VAEs is not trained to fully reconstruct the data but to generate reconstructions that can be well-classified by the discriminator. By doing this divergence measure, we can evaluate the goodness of the abstract representations for the given query. To compute the rate-distortion function, we trained several VAEs using different $\beta$ to study the rate-distortion trade-off in different regimes and the potential differences between vanilla VAEs and our model.

Using this model we can investigate whether the latent space can efficiently encode just the relevant invariances and symmetries required for the downstream task without the need to generate faithful reconstructions of the data. If that is the case, full reconstruction no longer becomes a necessary condition for goal-oriented representations. In the next section, we present the main results and their connection to the theoretical framework presented previously.

\subsection{Results}
\begin{figure}
    \begin{subfigure}[h]{0.3\linewidth}
    \includegraphics[width=1\linewidth]{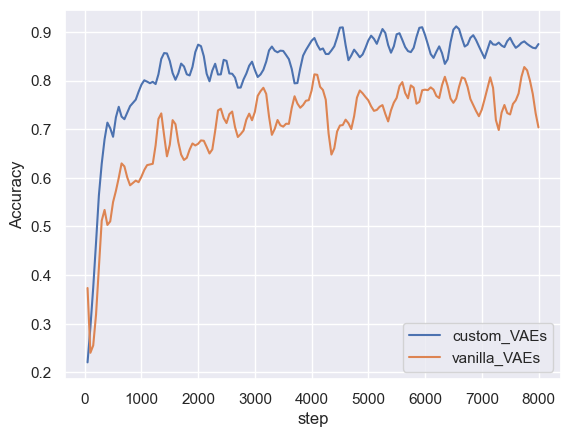}
    \caption{Accuracy as a function of the number of steps.}\label{acc_steps}
    \end{subfigure}
    \hfill
    \begin{subfigure}[h]{0.3\linewidth}
    \includegraphics[width=1\linewidth]{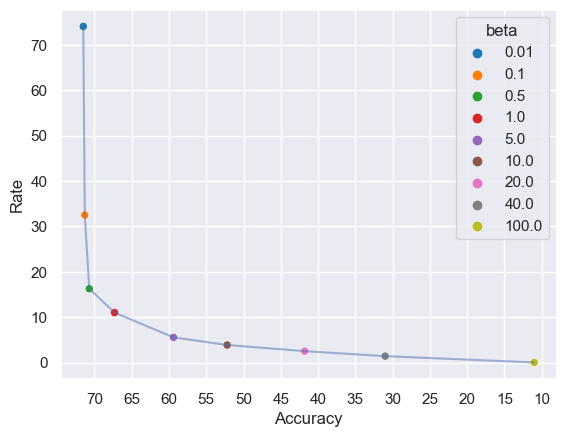}
    \caption{Fully reconstruction objective.}\label{fully_rec}
    \end{subfigure}
    \hfill
    \begin{subfigure}[h]{0.3\linewidth}
    \includegraphics[width=1\linewidth]{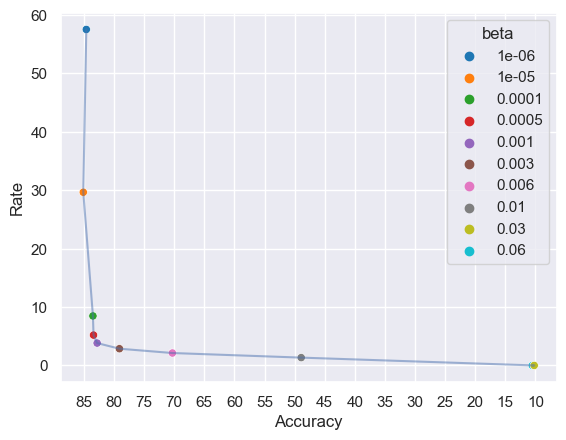}
    \caption{Action-centric objective.}\label{abs_rec}
    \end{subfigure}%
    \caption{Rate-distortion curve made up of different VAEs. Note that as a distortion measure here we use the accuracy. Left figure: vanilla VAEs that try to maximize the mutual information given the channel constraints. Right figure: lossy compression VAEs whose function is to maximize utility downstream given the channel constraints.}\label{rd_curve_vaes}
\end{figure}
The results regarding the transmission of information in the two different VAEs are shown in Figure \ref{rd_curve_vaes}. In (Figure \ref{acc_steps}) it can be seen how action-oriented VAEs converges faster to an encoding-decoding scheme that is useful for the downstream task (measured by the accuracy), compared to the VAEs. This indicates that action-centric representations might require less exposure to data, which makes them more efficient in terms of exploiting the available information.

Figure \ref{fully_rec} and Figure \ref{abs_rec} show the rate-distortion curve for both types of VAEs. It is clear how action-oriented representations require significantly less information from the data to achieve better results in the downstream task. In particular, it can be seen that transmitting at a rate of around $10$ bits the action-centric VAEs reaches almost $85\%$ of accuracy, compared to the $67\%$ achieved by the vanilla VAEs at approximately the same rate. This suggests that lossy compression leads to efficient codings and, importantly, to better behaviour.

\begin{figure}
    \centering 
    \hspace*{0cm} 
    \includegraphics[width=1\textwidth]{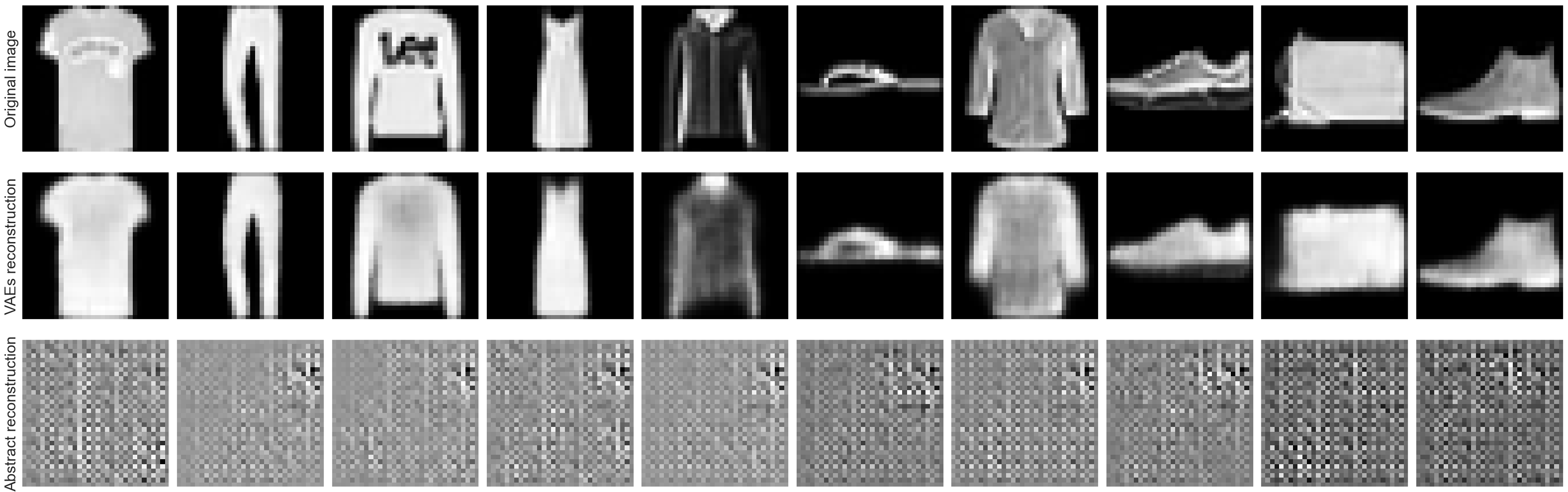}
    \caption{Data reconstruction by VAEs and action-centric VAEs.}\label{fig:reconstructions}
\end{figure}

The results so far indicate that the main function of representations might not be to fully reconstruct the data, but not capture the relevant invariances in the data exploited by optimal behaviour. We explicitly show this by investigating the reconstructions obtained by action-oriented representations. Figure \ref{fig:reconstructions} shows a sample of the reconstructions obtained by the vanilla and action-centric VAEs, respectively. While the vanilla VAEs generates relatively faithful reconstructions of the data, the action-centric VAEs generates meaningless and uninterpretable images. Interestingly, these action-oriented reconstructions are classified with approximately $85\%$ of accuracy, which suggests that the underlying structure of these reconstructions is preserving some important invariances and symmetries of the data. On the contrary, the full reconstruction might carry irrelevant information that is non-task specific, which could explain why they are more difficult to classify. 

\raggedbottom
\section{Discussion}
Agents need to navigate complex environments with limited biological information processing. Under this circumstance, an optimal perceptual system has to efficiently allocate cognitive resources to transmit the relevant sensory information to achieve successful behaviour. Thus, the goal of perception is not to generate faithful reconstructions of the sensory input, but abstract representations that are useful downstream.

A common approach to representations in Artificial Intelligence and Neuroscience is that they should be in service of fully reconstructing the data. However, such representations will carry irrelevant information for downstream tasks that only depend on the exploitation of specific invariances and symmetries of the data.

In this work, we explore useful efficient coding within the framework of rate-distortion to explore optimal information processing for task-dependent contexts. We have provided a formal definition of abstractions that can be used to learn action-centric representations whose main function is to capture the task-dependant invariances in the data. Such lossy compressions of the data lie near optimal points of the rate-distortion curve. Crucially, we show that action-centric representations i) are efficient lossy compressions of the data; ii) capture the task-dependent invariances necessary to achieve adaptive behaviour; and iii) are not in service of reconstructing the data. This could shed some light on how organisms are not optimized to reconstruct their environment; instead, their representational system is tuned to convey action-relevant information.

Interestingly, our work resonates with recent research on multimodal learning such as the joint embedding predictive architecture and multiview systems \cite{lecun_vicreg,robust_repr}. The main objective of these models is to obtain representations that are useful downstream but from which it's not possible to reconstruct the data. These representations learn the relevant invariances by maximizing only the information shared across different views or modalities of the data.  We argue that action-centric representations operate in a similar way, as shared information across views is an implicit way to define a query (see Appendix \ref{multiview_app}).

An interesting line of research is to explore faithful reconstruction in the context of fine-grained queries such as pixel predictability. We hypothesize that, as the number of pixel-specific queries approaches the pixel space of the image, the abstract representation might allow for faithful reconstruction of the data. Although that could be to the detriment of worse performance on downstream tasks.

In conclusion, this work sets a promising line of research in the field of representational theory by understanding representations not as faithful reconstructions of the data but as action-driven entities.

%
%
%
\bibliographystyle{splncs04}
\bibliography{references}

\newpage
\appendix
\section{Model details}
The classifier used to implement the query is a deep convolutional network (CNN) with three convolutional layers. The number of filters for the first layer is 16, and it is doubled in each layer. The kernel size is 3 in all layers, and padding is set to 1, also in all layers. Stride is 1 in the first two layers, and 2 in the third one. In addition, batch normalization is applied in each layer; 16 for the first one, and doubled in each layer. The activation function in each layer is ReLU, and max pooling is applied in the first two layers, both with a kernel size of 2, and stride of 2 in the first and 1 in the second. Between the first two fully connected layers it is used a dropout of 0.2. The number of neurons for the fully connected layers is 512, 128, and 10. We use the Adam optimizer with a learning rate of 0.001. We trained the classifier for 15 epochs with a batch size of 64.  

Regarding the VAEs, the encoder is a CNN of 4 layers with the same parameters as the CNN. Every VAEs trained has 8 latent dimensions and are trained for 20 epochs using a batch size of 64. In the case of the vanilla VAEs, the $\beta$ used to draw the rate-distortion curve are 100, 40, 20, 10, 5, 1, 0.5, 0.1, and 0.01. For the custom VAEs, the $\beta$ values are 6e-2, 3e-2, 1e-2, 6e-3, 3e-3, 1e-3, 5e-4, 1e-4, 1e-5, 1e-6.

\section{Latent space of VAEs}
PCA to explore and show the latent space of the vanilla and action-centric VAEs that achieve a good performance downstream:
\begin{figure}
    \begin{subfigure}[h]{0.5\linewidth}
    \includegraphics[width=1\linewidth]{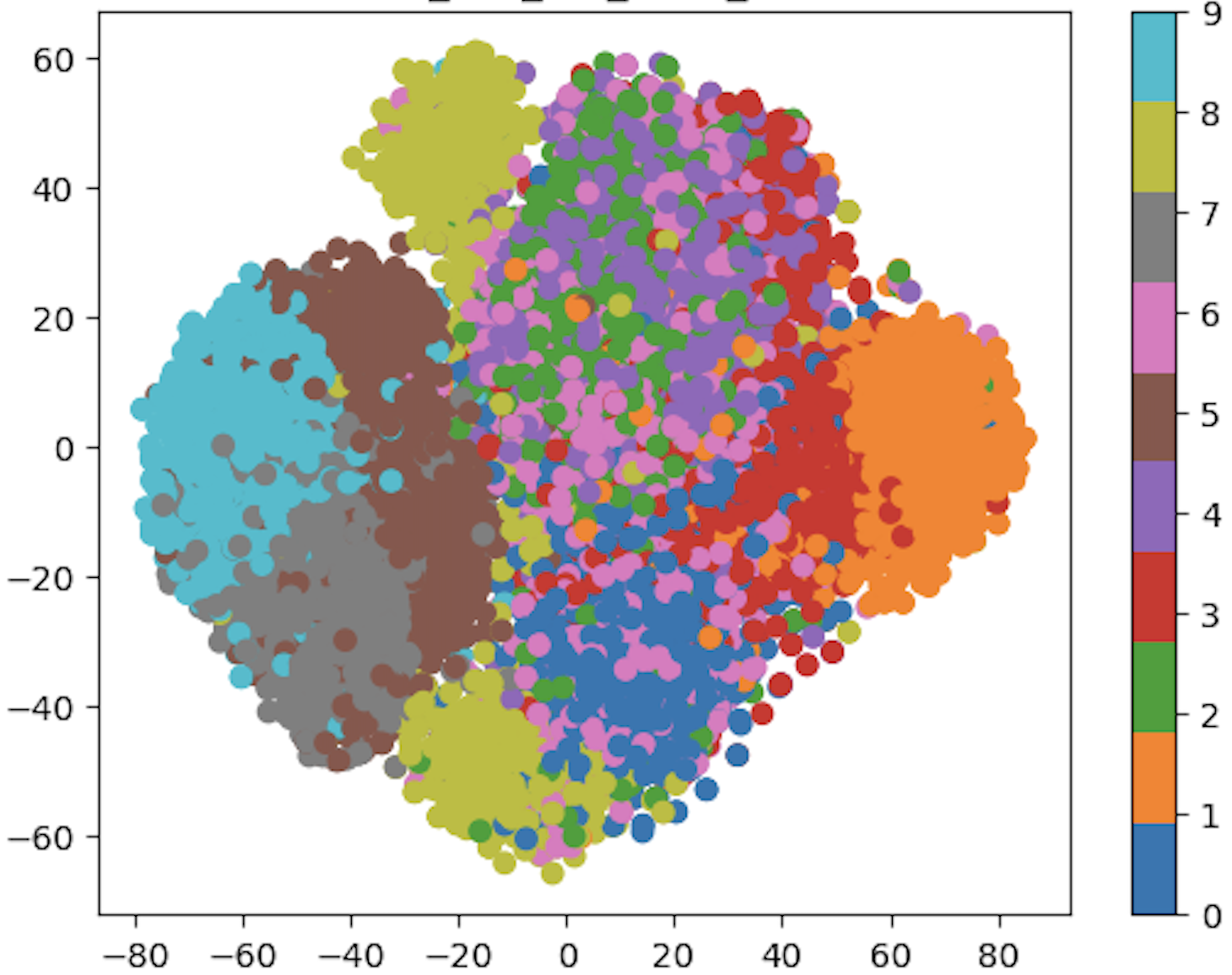}
    \caption{Vanilla VAEs ($\beta = 0.5$)}
    \end{subfigure}
    \hfill
    \begin{subfigure}[h]{0.5\linewidth}
    \includegraphics[width=1\linewidth]{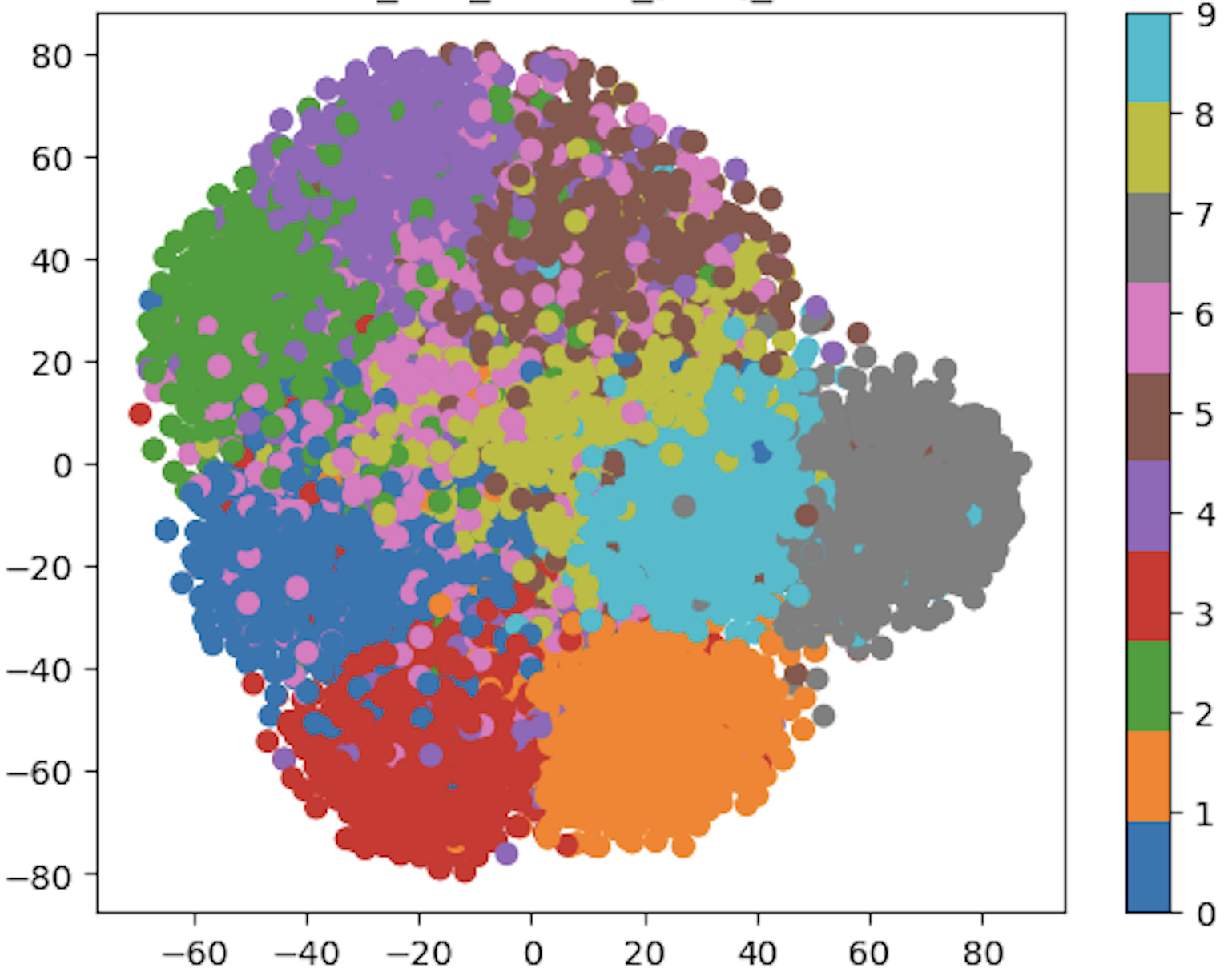}
    \caption{Action-centric VAEs ($\beta = 1e-4$)}
    \end{subfigure}%
    \caption{Latent spaces of the two types of VAEs. }
\end{figure}

As can be seen, our VAEs achieves a compact meaningful encoding of the data, with an apparent better separability among classes than the vanilla VAEs.

\section{ELBO and RDT}\label{elbo_mutual}
One way to derive the upper bound on mutual information from the complexity term of the ELBO is:
\begin{align}
    \mathop{\mathbb{E}}_{p(x)} \bigl[ D_{KL}[q(z|x) || p(z)] \bigl] &= \mathop{\mathbb{E}}_{p(x)} \bigl[ \int q(z|x) \ln{\frac{q(z|x)}{p(z)}}dxdz \bigl]\\
    &= \mathop{\mathbb{E}}_{p(x)} \bigl[ \int q(z|x) \ln{\frac{q(z|x)q(z)}{p(z)q(z)}}dxdz \bigl]\\  &= \mathop{\mathbb{E}}_{p(x)} \bigl[ \int q(z|x) \ln{\frac{q(z|x)}{q(z)}}dxdz +  \int q(z|x) \ln{\frac{q(z)}{p(z)}}dxdz \bigl]\\
    &= \int q(z|x)q(x) \ln{\frac{q(z|x)}{q(z)}}dxdz + \int q(z|x)q(x) \ln{\frac{q(z)}{p(z)}}dxdz\\
    &= \int q(x,z) \ln{\frac{q(x,z)}{q(x)q(z)}}dxdz + \int q(z) \ln{\frac{q(z)}{p(z)}}dz\\
    &= I(X;Z) + D_{KL}[q(z)||p(z)]\\
    &\geq I(X;Z)
\end{align}

Another way to derive this upper bound is by splitting the expected complexity into conditional entropy and entropy terms:

\begin{align}
    \mathop{\mathbb{E}}_{p(x)} \bigl[ D_{KL}[q(z|x) || p(z)] \bigl] &= \mathop{\mathbb{E}}_{p(x)} \bigl[ \int q(z|x) \ln{\frac{q(z|x)}{p(z)}}dxdz \bigl]\\
    &= \int q(z|x)q(x) \ln{q(z|x)}dxdz -  \int q(z|x)q(x) \ln{p(z)}dxdz\\
    &= \int q(x,z) \ln{q(z|x)}dxdz - \int q(z) \ln{p(z)}dz \label{last_exp}
\end{align}

In the last equation, we can see that the first term is the negative conditional entropy $-H(Z|X)$ which is one of the two terms in which the mutual information is decomposed: $I(Z;X) = H(Z) - H(Z|X)$. To get the entropy $H(Z)$ we need to replace $p(z)$ by an approximate distribution $q(z)$. By Jensen's inequality, we know that $D_{KL}[q(z)||p(z)] \geq 0$, therefore, we know that:

\begin{align}
    \int q(z) \ln{q(z)} - \int q(z) \ln{p(z)} &\geq 0\\
    \int q(z) \ln{q(z)} &\geq \int q(z) \ln{p(z)}
\end{align}

Replacing that term in the previous expression 
\eqref{last_exp} we get:

\begin{align}
    \int q(x,z) \ln{q(z|x)}dxdz - \int q(z) \ln{p(z)}dz\ &\geq \int q(x,z) \ln{q(z|x)}dxdz - \int q(z) \ln{q(z)}dz\\ 
    \int q(x,z) \ln{q(z|x)}dxdz - \int q(z,x) \ln{q(z)}dxdz
    &\geq \int q(x,z) \ln{\frac{q(x,z)}{q(x)q(z)}}dxdz = I(X;Z)
\end{align}

\section{Multiview architecures and queries}\label{multiview_app}

Given a query $Q(X)$ over $X$ in a multiview scenario it can be understood as the subset of information contained in the intersection of $X$ and $t(X)$ such that:

\begin{align}
    Q(X) \in X \cap t(X)
\end{align}

as the transformation $t$ only preserves those symmetries relevant for the query (i.e., relevant to solve a set of tasks that only depend on those invariances). Therefore, the relevant query in a multiview scenario can be defined as:

\begin{align}
    Q(X) = p(X, t(X))
\end{align} 

Mutual information between $X$ and $Z$ and between $Q(X)$ and $Z$ is (assuming that X, X' and Z form a dag where Z only depends on X):
\begin{align}
    I(X;Z) &= \int p(x,z) \ln{\frac{p(x,z)}{p(x)p(z)}}dxdz\\
    I(Q(X);Z) &= \int p(q,z) \ln{\frac{p(q,z)}{p(q)p(z)}}dqdz\\
    &= \int p(x,x',z) \ln{\frac{p(x,x',z)}{p(x,x')p(z)}}dxdx'dz\\
    &= \int p(x,x',z) \ln{\frac{p(x')p(x|x')p(z|x)}{p(x')p(x|x')p(z)}}dxdx'dz\\
    &= \int p(x,x',z) \ln{\frac{p(z|x)}{p(z)}}dxdx'dz\\
    &= \int p(x,x')p(z|x) \ln{\frac{p(x,z)}{p(x)p(z)}}dxdx'dz\\
    &= \int p(x)\frac{p(x,z)}{p(x)} \ln{\frac{p(x,z)}{p(x)p(z)}}dxdz\\
    &= \int p(x,z) \ln{\frac{p(x,z)}{p(x)p(z)}}dxdz\\ &= I(X;Z) = I(X;X')
\end{align}

The mutual information between the latent $Z$ and one of the views $X$ is equal to the mutual information between the query distribution $Q(X)$ and the latent $Z$. As the mutual information between an optimal lossy representation and its corresponding view is equal to the mutual information between views, then, the information conveyed by the query is the one shared by the views. This shows that the multiview architecture is essentially a query-oriented system where the transformations applied to the data keep specific invariances with respect to a set of implicit queries of interest.
\end{document}